\documentclass[manuscript,screen,authorversion]{acmart}

\AtBeginDocument{%
  \providecommand\BibTeX{{%
    \normalfont B\kern-0.5em{\scshape i\kern-0.25em b}\kern-0.8em\TeX}}}

\setcopyright{acmcopyright}
\copyrightyear{2022}
\acmYear{2022}

\usepackage{bbm,xspace}

\newcommand\hull{$\mathcal{H}^{tr}$\xspace}
\newcommand\dtr{$\mathcal{D}^{tr}$\xspace}

\begin{document}

\title{To what extent should we trust AI models when they extrapolate?}

\author{Roozbeh Yousefzadeh}
\email{roozbeh.yousefzadeh@yale.edu}
\affiliation{%
  \institution{Yale Center for Medical Informatics and VA CT Healthcare System}
  \city{New Haven}
  \state{CT}
  \country{USA}
  \postcode{06511}
}

\author{Xuenan Cao}
\affiliation{%
  \institution{MacMillan Center for International and Area Studies, Yale University}
  \city{New Haven}
  \state{CT}
  \postcode{06511}
  \country{USA}}
\email{xuenan.cao@yale.edu}


\begin{abstract}
Many applications affecting human lives rely on models that have come to be known under the umbrella of machine learning and artificial intelligence. These AI models are usually complicated mathematical functions that make decisions and predictions by mapping from an input space to an output space. Stakeholders are interested to know the rationales behind models' decisions; that understanding requires knowledge about models' functional behavior. We study this functional behavior in relation to the data used to create the models. On this topic, scholars have often assumed that models do not extrapolate, i.e., they learn from their training samples and process new input by interpolation. This assumption is questionable: we show that models extrapolate frequently; the extent of extrapolation varies and can be socially consequential. We demonstrate that extrapolation happens for a substantial portion of datasets more than one would consider reasonable. How can we trust models if we do not know whether they are extrapolating? Given a model trained to recommend clinical procedures for patients, can we trust the recommendation when the model considers a patient older or younger than all the samples in the training set? If the training set is mostly Whites, how do we measure the extent to which we can trust its recommendations about Black and Hispanic patients? How do we know if extrapolation is significant for a given sample if we do not compare the sample to the training data? Which dimension (race, gender, or age) does extrapolation happen? Even if a model is trained on people of all races, it still may extrapolate in significant ways related to race. So the leading question is, to what extent can we trust AI models when they process inputs that fall outside their training set? This paper investigates several social applications of AI, showing how models extrapolate without notice. The difficulty of auditing is that datasets have many features, thus hard to review individually. To solve this problem, we use a systematic method of geometric analysis. We also look at different sub-spaces of extrapolation for specific individuals subject to AI models and report how these extrapolations can be interpreted, not mathematically, but from a humanistic point of view. We recommend that AI pipelines include this auditing approach. If extrapolation is statistically excessive, the model's output should be sent for review by human experts and provide information about the extent and dimensions of extrapolation.
\end{abstract}




\maketitle

\section{Introduction}

Can we trust the decisions of AI models? How do we know if we can trust it? To answer these questions, we need to know (1) the specific tasks for which a model is intended, and (2) the data on which it has been trained. In this paper, we focus on these two aspects with illustrative examples in the results section. 

There has been a considerable effort to promote accountability and transparency in the use of machine learning models. Much research has focused on developing interpretable models for various tasks. Another branch of study has focused on developing methods to demystify black-box models that are hard to interpret. Some researchers and policymakers have proposed that any model output should be accompanied by an explanation understandable to the person affected. Stakeholders should be able to understand what goes into an AI decision. Regulations informed by these suggestions have recently started taking effect. For instance, the European Union has drafted and passed the General Data Protection Regulation (GDPR) \citep{voigt2017eu}. The regulation responds to cases in which people were deprived of any explanation about consequential decisions made by the models for them. An example of such deprivation is in recidivism prediction used for bail decisions \citep{rudin2018age}. 
Interpretability is helpful to the concerned party as well as policymakers and scholars attempting to map and measure models' attitudes towards demographic groups to expose and overcome their biases and shortcomings \citep{yousefzadeh2020auditing}. 

In this paper, we show that when making decisions or predictions about new samples, machine learning models often have to extrapolate outside the samples on which they have been trained. Extrapolation refers to the type of estimation that goes beyond the convex hull of training set. For example, a model that has learned to predict the likelihood of a person defaulting on their next credit card payment may often need to project outside the original observation range provided by the training set. For researchers interested in the social implications of extrapolation, we provide three case studies in the results section to translate the math problem into a human problem. In translating the math problem, we provide accessible examples that require no specialized knowledge to understand. 

The extent of extrapolation is crucial information about the model's decision. We recommend the extent of extrapolation be reported in favor of accountability and transparency. We propose that AI models automatically log their extrapolation.

Mathematically, extrapolation is a concept defined in contrast to interpolation. A given model, $\mathcal{M}$, is trained on a finite set of samples, i.e., a training set. Future decisions/predictions of a model will be based on what it has learned from its training samples. This automatic learning process has come to be known as artificial intelligence (AI) and machine learning. Moreover, because learning is automatic via complex mathematical procedures, the resulting models are often hard to interpret, making them black-box models.

A training set, however large, will have a finite set of samples, and those samples form a convex hull. Here, we denote the training set by \dtr and the convex hull of the training set by \hull. Eventually, $\mathcal{M}$ will be used to make predictions/decisions about samples other than the ones in \dtr. These other samples are referred to as testing samples. Whenever $\mathcal{M}$ makes a decision for a new sample $x$, that process is either interpolation or extrapolation. If $x$ is inside the convex hull of \dtr, then the operation performed by $\mathcal{M}$ is an interpolation; otherwise, it is called extrapolation. A convex hull is the smallest interval containing all the data abscissae \citep{ascher2011first}.

Consider, for example, a case where a model is trained to make clinical decisions for patients. The model would be trained on data from previous patients gathered in a training set. Now, assume that patients' age in the training set varies between 25 and 60. When the model receives information about a patient younger than 25 or older than 60, it would need to extrapolate outside the convex hull of its training set. This seems straightforward, but extrapolation may happen even for patients aged between 25 and 60 when we consider other features. For example, the age of samples in the training set might vary between 30 and 40 for Hispanics. When the model decides for a 50 years old Hispanic patient, it would still be extrapolating because 50-year-old for a Hispanic patient falls outside the convex hull of the training set. What if we have a whole host of other variables about other demographic features, blood measurements, other lab tests, height, weight, and other patient information? For patients with many features, extrapolation may occur in many obscure ways not readily evident to physicians and model users.


\subsection{Our contributions}

Our contributions can be summarized as the following:

\begin{enumerate}
    \item We show that for several standard datasets used in social applications of AI, a considerable portion of testing samples fall outside the convex hull of the training set. Evidently, the functional task of machine learning models often involves extrapolation.
    \item We show that the extent of extrapolation for testing samples is not negligible. We further analyze the patterns of the required extrapolations.
    \item We propose a new requirement for achieving the explainability of AI models: any extrapolation performed by AI models should be reported to the relevant parties affected by the outcome. For example, when a model used for medical decision making has to extrapolate to decide for a new patient, the model should report, to the physician in charge, the fact that the new patient falls outside the convex hull of the data it has been trained on, namely, the model has extrapolated.
\end{enumerate}


\subsection{Our plan}

First, we review the literature on the accountability and transparency of AI models to show the current roadblocks that we will help clear. We draw on the literature of extrapolation and the geometry of datasets. Second, in Section~\ref{sec:formulation}, we present the formulation for investigating the extent of extrapolation for any testing sample. Our systematic approach is model agnostic because it promotes transparency by analyzing the geometry of data. In our formulation, we consider domains with a combination of categorical, integer, and continuous variables. In Section~\ref{sec:results}, we present our results on several datasets widely used in research. Most research published on some of these datasets has discussed bias inherent in the data, but few offer ways to audit how much the models extrapolate. We propose a solution. When extrapolation happens, the model should have a self-auditing system to report the extent of extrapolation. In section~\ref{sec:policy}, we discuss the policy implications of our work, followed then by the conclusion.

\section{Literature review} \label{sec:litreview}

\subsection{Studies on promoting accountability and transparency}

The past few years have seen promising progress in promoting accountability and transparency in using AI models. Some have made headway in developing inherently interpretable models. Such models need to be designed and tailored for individual tasks, e.g., financial lending \citep{chen2022holistic}, embryo selection \citep{afnan2021interpretable}, recidivism prediction \citep{zeng2017interpretable}, etc. This judicious use of models is advantageous, especially for high-stakes decision-making. For such applications, the common wisdom has been to avoid black-box models because they are hard to interpret. Many interpretation methods have failed to explain the rationale behind decisions of AI models reliably. In fact, some widely used interpretation methods such as LIME \citep{ribeiro2016should} may even produce explanations that contradict the models' decisions. Because of that, \citet{rudin2019stop} and others have suggested that black-box models should not be used in high-stakes applications. In support of this approach, studies show that in some learning tasks, deep networks and other black-box models do not have an accuracy advantage over simple models, especially if one performs a modest pre-processing on the data \citep{rudin2021interpretable}.

However, one appeal of machine learning, and more often of deep learning, is automation (i.e., learning from data without feature engineering or sophisticated model design). It is known that neural networks are universal approximators \citep{cybenko1989approximation,strang2019linear}, meaning that they can approximate any function over a domain with bounded error. As a result, for many learning tasks, black-box models such as neural networks are frequently used to avoid feature engineering, or else prevent transparency and keep the models proprietary \citep{rudin2019why}. In the meantime,  better methods have been developed for interpreting, auditing, and debugging black-box models \citep{yousefzadeh2020deep}. Some of the proposed interpretation methods are model agnostic \citep{yousefzadeh2020auditing}. Some others are specific to certain model types, e.g., decision tree models \citep{lundberg2020local}. There are also studies focused on making the explanations better understandable by non-experts \citep{wachter2018counterfactual,shneiderman2020bridging} using counterfactuals and interactive user interfaces.

One of the motivations for using machine learning models is that these models can learn by extracting hidden patterns and structures in the data. Based on that learning, they can generalize and make decisions and predictions about unseen data with reasonable accuracy, sometimes more than humans. If we agree that a model has learned from a dataset, should we expect that the model will generalize well on any related data? What if the model has to extrapolate outside the convex hull of its training data? Studies in psychology and cognitive science show that humans and machines are usually less confident in their decisions and predictions when they extrapolate instead of interpolating, and a correlation exists between human confidence and the correctness of predictions \citep{stojic2018you}.

We need computational methods and procedures to interpret the models to promote accountability and transparency. It is also important to mandate auditing for black-box models and explanations for outputs. In that spirit, rules and regulations have been proposed such that decisions made by AI models be accompanied with clear explanations for individuals affected by decisions \citep{wachter2018reasonable}. One culmination of the many years of advocacy is the European Union's GDPR. Other organizations, including the World Health Organization (WHO), have adopted guidelines for ethics and governance of AI in their field \citep{world2021ethics}.

Building on these endeavors, we suggest that the geometry of data should also be part of the routine procedures for auditing and explaining AI models. The geometry of data is an aspect of data that has received less attention, especially in instances where models have to extrapolate outside the convex hull of their training set. The formulation section and the results section demonstrate this point. There, we show that AI models extrapolate frequently and significantly outside the range of what they have encountered in the training process. We believe that routine and automatic reporting about the extent of extrapolation can promote transparency and accountability.

\subsection{Studies on extrapolation and geometry of datasets}

Geometry is one important aspect of data which also intimately relates to statistics and other mathematical properties. Many studies in the AI literature have focused on the geometric understanding of data, e.g., \citep{cohen2020separability,lebanon2005riemannian}. Geometry helps understand models' functional behavior and distinguish whether that involves interpolation or extrapolation. This approach provides clues to the functional operation of a model in relation to the training data. By definition, when a function, $\mathcal{M}$, operates on input outside its convex hull of training set, \hull, it extrapolates. In other words, a function has to extend outside its \hull to reach an input and process it.

The distinction between interpolation and extrapolation has long been part of machine learning literature. For example, \citet{haffner2002escaping} developed extrapolating support vector machines (SVMs) to classify datasets such as MNIST. \citet{vincent2002k} used the distance to convex hulls for classification with K-Nearest Neighbor algorithms. Extrapolation also has a rich literature in many fields of study, including pure and applied mathematics\citep{fefferman2005interpolation,sidi2003practical}, cognitive science \citep{griffiths2008modeling}, psychology \citep{silliman2021extrapolation}, neuroscience \citep{guigon2002neural}, etc. Mathematicians have studied how a learned function can be extended in certain ways outside a domain, e.g., Whitney's extension problem \citep{fefferman2006whitney}. \citet{yousefzadeh2021extrapolation} recently reviewed the cognitive psychology literature that studies extrapolation and learning.

However, extrapolation has been neglected in recent literature on the mathematical understanding of black-box models and in the scholarship on accountability and transparency of AI models. In recent literature, extrapolation usually implies using a trained model for a task different from the task it has been trained on. For a model trained for object recognition, extrapolation may imply using it to classify hand-written digits or radiology images of the liver, i.e., out-of-distribution samples. With that thinking, \citet{xu2020neural} concluded that neural networks are not good at extrapolation.

Indeed, there is a connection between extrapolation and the topic of out-of-distribution detection \citep{krueger2021out}. But extrapolation does not only happen for out-of-distribution samples. While a model may need to extrapolate to process an input, being outside the \hull does not necessarily imply that input is out-of-distribution. Let us be clear: we do not imply that all samples outside the convex hull of the training set should be considered out-of-distribution. Extrapolation and learning frequently supplement each other, but evidence suggests that both humans and AI models are more susceptible to mistakes when they extrapolate as opposed to when they interpolate \citep{busemeyer1997learning,stojic2018you,sharma2009scale,webb2020learning}. This is sometimes reported as extrapolation tasks being more difficult to perform correctly \citep{griffiths2008modeling,delosh1997extrapolation,mcdaniel2005conceptual}. Given this, we caution that one should be informed, and perhaps be more cautious and less confident, about the instances where AI models extrapolate.

Some studies assume that machine learning models generalize merely by interpolating between training samples \citep{webb2020learning}, but geometric evidence does not support that presupposition. Other studies assume that learning is solely about interpolation \citep{belkin2019reconciling}. Yet, this presumption neither explains how machine learning models generalize, nor aligns with human understanding of learning. Recently, \citet{balestriero2021learning} concludes that ``learning almost always amounts to extrapolation" in any high dimensional space (more than 100 dimensions). We show that even in datasets with as few as 23 dimensions and more than a million samples, extrapolation is not negligible for a considerable portion of samples. This is the third case study in the results section. Therefore, the number of dimensions might not be a good guide to determine whether a model is interpolating or extrapolating. Instead, for any given sample, one should solve an optimization problem (described in Section~\ref{sec:formulation}) to verify whether the sample is inside or outside the \hull. If the sample is outside the \hull, one should examine the directions of extrapolation.

All the above literature review amounts to our suggestion that interpretation frameworks should report if the model has extrapolated for each testing sample, the features involved in extrapolation, and the extent of extrapolation in those features. With such information, an expert and the affected parties will stay informed. We further suggest that directions of extrapolation be investigated systematically, using advanced techniques in numerical linear algebra \citep{golub2012matrix} to identify the dominant patterns and most influential features.



\section{Formulation} \label{sec:formulation}

Our trained model, $\mathcal{M}$, is a function that maps inputs to outputs. It follows that $\mathcal{M}$ has a domain and a range. The domain of the model may be bounded or unbounded depending on the nature of the variables. For example, for a variable representing the age of people, the domain will be lower-bounded by zero. It is also sensible to assume that the age variable is upper-bounded by 120 or so to represent the cap of life expectancy. The domain may be discrete for certain variables, such as a categorical variable that represents the occupation of people. We denote the domain of $\mathcal{M}$ by $\Omega$. $\Omega$ could be comprised of discrete and continuous spaces.

\subsection{Convex hull and the domain}

$\mathcal{M}$ is typically trained on a training set. By definition, training samples as well as testing samples are all contained in the domain, i.e., $\mathcal{D}^{tr} \in \Omega$. The convex hull of the training set, by definition, contains any convex combination of samples in \dtr. As a result, the convex hull of samples with categorical features may contain points that do not conform to the categories. 

For example, consider two samples in \dtr representing two patients, one female and one male. Patient information may include their blood test measurements. A convex combination of these two patients could be points that are neither male nor female, rather it could be 50\% female and 50\% male. Similarly, we can consider three training samples for a dataset about people's salaries and occupations: a nurse, a farmer, and a programmer. Convex combination of these samples could be a point representing someone who is 20\% nurse, 50\% farmer, and 30\% programmer. These convex combinations might not always be meaningful for certain types of studies. It is not hard to imagine someone being a nurse, a farmer, and a programmer simultaneously, but that might not be representative of samples we have in \dtr. In fact, we might have extracted some datasets from the census where each sample has only one occupation. In such a case, it could be more sensible to consider points in the convex hull that only have a single occupation. We will explain how such points may be computed. 

For continuous features, however, we may have different thinking. For example, if we have three samples, all nurses, with varying ages of 30, 40, and 50 years old, a convex combination of them could be a sample whose occupation is nurse, with age somewhere between 30 and 50. Therefore, we may only be interested in certain convex combinations, representing points with discrete values for certain categorical variables such as occupation. Conversely, sometimes we might want to relax that requirement for some categorical features and consider mixed categorical variables. This would be reflected in how we define the $\Omega$.

The discussion above signifies that all points in \hull will be within the bounds of $\Omega$, but they may not be a member of it when $\Omega$ is discrete in certain dimensions. Therefore, we would be interested in points that belong to the intersection of \hull and $\Omega$. Note that we may want to relax some or all of the discrete variables to study different dimensions of extrapolation. Perhaps, we want to relax the gender variable to include convex combinations of female and male samples. Moreover, for a female testing sample, we may want to compute the closest convex combination of males to her. We can do so by modifying the $\Omega$. In the next section, we formulate an optimization problem that can solve these questions.

\subsection{Projecting a query point to a convex hull}

Given a point $x$, in the feature space $\Omega$, we would like to find the closest point to it on the $\mathcal{H}^{tr}$. Distance can be measured using any desired norm. Here, we use the 2-norm distance and minimize it via the objective function
\begin{equation} \label{eq:hull_obj}
	\min_{x^h} \| x^h - x \|
\end{equation}
Our constraint, in a broad description, is that the solution lies within the intersection of $\mathcal{H}^{tr}$ and $\Omega$
\begin{equation} \label{eq:hull_c_main}
	x^h \in \mathcal{H}^{tr} \cap \Omega.
\end{equation}

When $\Omega$ represents a continuous space, $\mathcal{H}^{tr} \cap \Omega$ will be equivalent to $\mathcal{H}^{tr}$, but when $\Omega$ includes discrete variables, their intersection may only be a subset of each. To implement constraint~\eqref{eq:hull_c_main}, we break it into several smaller constraints, some of them ensuring that $x^h$ belongs to $\mathcal{H}^{tr}$, and other ensuring that $x^h$ belongs $\Omega$.

First, we relate $x^h$ to the samples in training set
\begin{equation} \label{eq:hull_c1}
  x^h = \alpha \mathcal{D},
\end{equation}
where $\mathcal{D}_\phi$ is the training set, formed as a matrix where rows represent $n$ samples and columns represent $d$ features. The next two constraints ensure that $x^h$ belongs to the convex hull of $\mathcal{D}$.
\begin{equation} \label{eq:hull_c2}
  \alpha  \mathbbm{1}_{n,1} = 1,
\end{equation}
\begin{equation} \label{eq:hull_c3}
  0 \leq \alpha,
\end{equation}
where $\mathbbm{1}$ denotes a column vector of ones.

Minimizing the objective function \eqref{eq:hull_obj} subject to constraints~\eqref{eq:hull_c1}-\eqref{eq:hull_c3} will lead to the point on $\mathcal{H}^{tr}$ closest to $x$. Additionally, when $\Omega$ has discrete features, we have to ensure that
\begin{equation} \label{eq:hull_c4}
    x^h \in \Omega.
\end{equation}
For example, if $\Omega$ includes a binary variable on whether patients have Hepatitis or not, we ensure that $x^h$ has a binary value for that variable. For categorical variables, such as primary occupation, we binarize the variable. Let us consider a dataset with $k$=20 types of occupation. We create k features in $x$, let us say its elements $x(1:k)$, each corresponding to one occupation type. Since primary occupation is a categorical variable, we may require that $x^h$ to have a binary value for each occupation type, $$\forall i \in \{1:k\} \; , \; x^h(i) \in \{0,1\}$$. Moreover, if each person has to have only one primary occupation or be unemployed, we may require that sum of these elements be at most one $$\sum_{i=1}^k x^h(i) \leq 1$$. These two constraints will ensure constraint~\eqref{eq:hull_c4} is satisfied for the occupation variable. If there are other categorical variables, we can impose more similar constraints.

Minimizing the objective function \eqref{eq:hull_obj} subject to constraints~\eqref{eq:hull_c1}-\eqref{eq:hull_c4} will then lead to the point on $\mathcal{H}^{tr} \cap \Omega$ closest to $x$. We denote this projection with a function

\begin{equation} \label{eq:hull_project}
  x^h = \mathcal{P}^h(x,\mathcal{H}^{tr},\Omega).
\end{equation}

From the optimization point of view, this is a constrained least squares problem \citep{golub2012matrix}. Since our optimization problem is convex, we are guaranteed to find its solution. There are fast algorithms to solve this problem efficiently \citep{nocedal2006numerical}, especially when $\Omega$ is continuous. In such cases, a Gradient Projection (GP) algorithm could be a good choice as it can rapidly change its active set of constraints. At each iteration of the GP algorithm, we first compute the Cauchy point (the feasible minimizer along the direction of derivatives), and then perform a subspace minimization for the samples that are not binding in constraint~\eqref{eq:hull_c3}. Gradient projection algorithm may be applied to the dual form of the problem as well, which could be computationally beneficial if the solution is composed of a large portion of samples in \dtr.

When $\Omega$ has categorical variables, it would be helpful to use discrete optimization algorithms in tandem with the gradient projection method. One approach we have successfully used is to first relax the discrete space into a continuous space, and then gradually impose the discrete requirement using a homotopy method. In separate work, we shall discuss these optimization algorithms. Our code is published online.


\section{Results} \label{sec:results}

The above statistical formulation shows how we can measure the extent of extrapolation through geometric analysis. In which dimensions and socially consequential ways do models project the query point (testing sample) onto the convex hull? This section measures the distance of projection in several social applications of AI to show how models extrapolate without our knowledge. We also provide narrative examples to illustrate the real-life impact of extrapolation.

In the following cases, extrapolation happens in multiple dimensions - both continuous features (such as a person's age and the months of trading, which continues on a spectrum) and categorical features (such as the country of origin and gender). A testing sample outside the convex hull requires the model to project in both categorical and continuous variables. The fact that the training set is finite necessitates extrapolation in areas against social conventions. In the following case studies, the extrapolation happens to the extent that both categorical and continuous variables have to change for the query point to reach the convex hull. 

The first case highlights extrapolations that are not negligible because it goes against our social norms related to gender, race, and social-economic status. The second case highlights the high frequency of extrapolation. The third case shows how models may need to extrapolate for a large clinical dataset. Together, these results showcase frequently used, widely studied, and high-stake datasets. By showcasing the non-negligible extrapolations, we make a crucial distinction between our contributions and the existing literature: the critical insight here is not that datasets contain biased materials. Instead, we emphasize that without our proposed systematic approach, models trained on such datasets may freely and significantly extrapolate without notice.

\subsection{The First Case: Adult Income Dataset}

Adult income dataset \citep{kohavi1996scaling} is a standard public dataset that predicts individuals' annual income based on census data. The categorical variables include race, gender, native country, education, marital status, work class, and occupation. Age, years of education, capital, and work hours per week are continuous features. 

We choose to study this dataset because, first, it is widely used to study social applications of AI \citep{Dua2017uci}, and second, the extent of extrapolation for this dataset has not been subject to research from the perspective of promoting transparency and accountability. Scholars interested in the social implications of income prediction will benefit from this analysis and the practical methods we suggest to investigate and report the extent of extrapolation.

\begin{figure}[h]
 \centering
 \includegraphics[width=.65\linewidth]{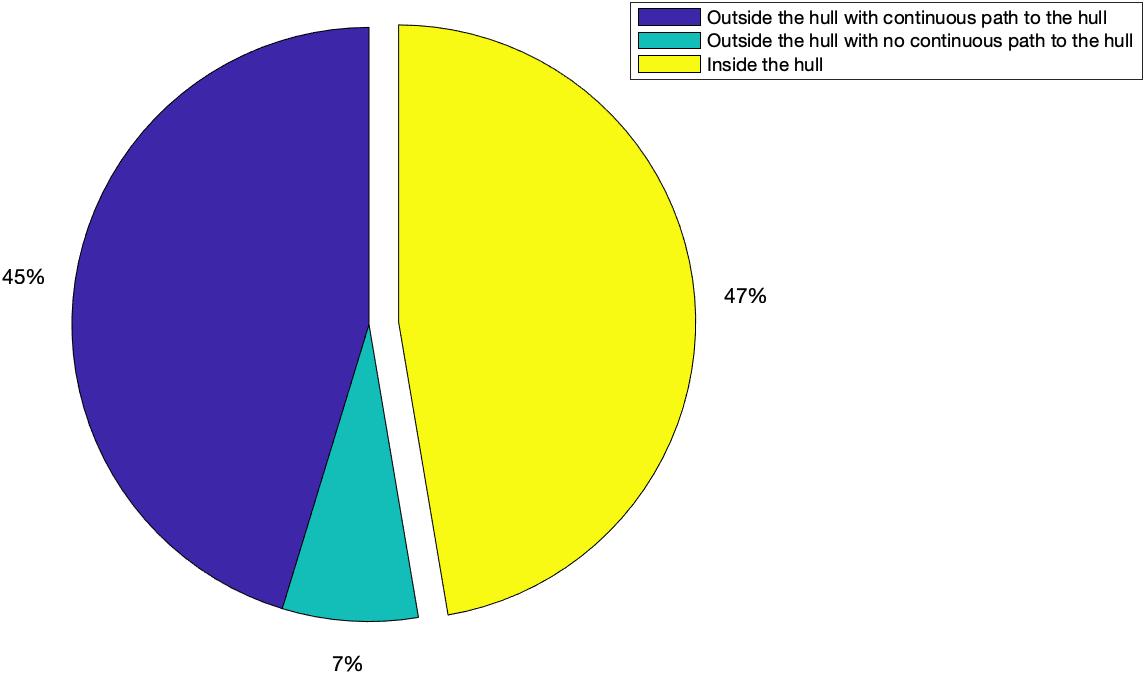}
 \caption{Pie chart showing the percentage of testing samples that are outside the convex hull of training set. For 7\% of those testing samples, there is not a single training sample with same gender, race, country of origin, marital status, and work class. As a result, any model would have to extrapolate along those categorical directions as well. For 45\% of testing samples, extrapolation happens in the dimensions of years of education, age, capital, and hours per week of work. For the remaining 47\% of testing samples, models may interpolate between the training samples. This statistic signifies that extrapolation is abundant and significant for testing samples of this dataset.}
 \label{fig:adult_pie}
\end{figure}

There are 16,281 testing samples and 32,561 training samples in this dataset. We see that 52\% of testing samples are considerably outside the convex hull of the training set, as shown in Figure~\ref{fig:adult_pie}. For 7\% of the samples, there is not even a continuous path to the \hull, meaning that for at least 1200 people, a model must extrapolate in categorical features and continuous features to reach the samples. In other words, for those testing samples, there is not a single training sample that has the same gender, race, country of origin, work class, and marital status.

\begin{figure}[h]
 \centering
 \includegraphics[width=.45\linewidth]{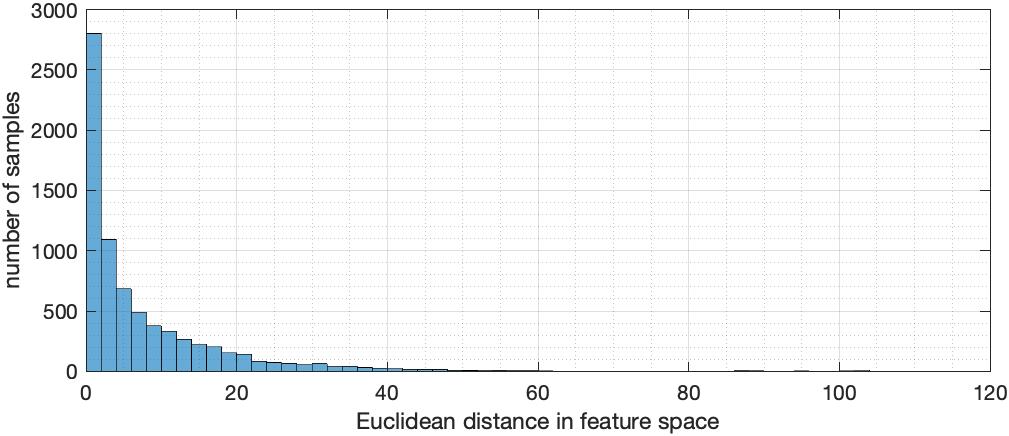}
 \caption{Distribution of distance to convex hull of training set for testing samples of Adult dataset that are outside the \hull with a continuous path to reach it. 52\% of testing samples for this dataset are outside the \hull.}
 \label{fig:adult_dist}
\end{figure}

Figure~\ref{fig:adult_dist} shows the distribution of distance to \hull for all testing samples that are outside the \hull with a continuous path to reach it, i.e., the 45\% of samples shown in Figure~\ref{fig:adult_pie}. As we can see, the extent of extrapolation is significant for most of these samples. For example, average rate of change is 29\% for years of education and 27\% for hours per week of work. This means that on average testing samples that are outside the \hull have to change their years of education by 29\%, and change their hours of work by 27\% in order to reach the \hull. These changes are not negligible for such people.

{\bf Directions to the \hull.} We further investigate the directions to \hull for all testing samples. Consider $\mathcal{D}^{te-out}$ to be the subset of testing set that is outside the \hull, formed as a matrix where rows represent samples, and columns represent features. We project these samples to the intersection of \hull and $\Omega$ using equation~\eqref{eq:hull_project} to obtain $$\mathcal{X}^h = \mathcal{P}^h(\mathcal{D}^{te-out},\mathcal{H}^{tr},\Omega)$$. Directions betweem the samples and the convex hull can be obtained via $$\mathcal{V} = \mathcal{X}^h - \mathcal{D}^{te-out}.$$

The matrix of directions, $\mathcal{V}$, has full rank, and its 2-norm condition number is $2 \times 10^{11}$. This condition number implies that there is some redundancy in the matrix of directions. Using rank-revealing QR factorization \citep{chan1987rank}, we identify that the workclass of ``Work without pay" is the redundant feature. Dropping this feature from the $\mathcal{V}$ will reduce its condition number to $4\times 10^3$. Singular value decomposition (SVD) of $\mathcal{V}$ indicates that there are 5 predominant patterns in directions. Analyzing the spectrum of directions indicates that there are 2 major clusters in $\mathcal{V}$.

{\bf Story Time: what happens to a person subject to extrapolation?} We now explain, in more detail, one of the testing samples that are outside the training set with no continuous path to the convex hull, i.e., one of the samples in the 7\% described in the pie chart in Figure~\ref{fig:adult_pie}. For illustration, we call this sample Anong. 

Anong is a 55-year-old unmarried woman, with no child, originally from Thailand and now working in the United States. She has a master's degree and 14 years of education and works in the private sector. She works 40 hours per week as an executive or a manager (i.e., in the Executive/Managerial work class). A machine learning system wants to guess, or perhaps decide, what her salary is or should be. A model (regardless of its type, such as boosted tree, neural network, support vector machine, etc) would be trained on a training set with 32,561 samples. But not a profile with the exact features of Anong exists in the training set. Anong's profile falls outside the convex hull of the training set, and any model trained on this dataset would need to extrapolate to reach her in the feature space. Let us investigate the extent of this extrapolation, not as a mathematical problem but as a human problem.

If there were a continuous path between the convex hull of the training set and Anong's information, the model would have extrapolated in the space of continuous features, a more straightforward task. But Anong's case is not as simple. To the model, Anong is a query point that requires extrapolation in multiple categorical and continuous features to project Anong's salary. 

One extrapolation path between Anong and the \hull is to change her work class. If we adjust her work class to Administrative/Clerical, the closest point on the convex hull can be represented as a 26-year-old single Thai woman with 13 years of education working in the private sector. Let us call this point Malee. Now, imagine if we have to present the resumes of Anong and Malee to HR personnel, they would be treated as entirely different candidates. The extent of extrapolation is significant in a socially understandable way. Their profiles are similar only because they are both Thai (the categorical variable of the country of origin), unmarried (the categorical variable of marital status), and woman (the categorical variable of gender). Although in the real world, the country of origin, marital status, and gender should not be the basis for determining a person's salary, these factors have been shown again and again to be good predictors for one's salary, and therefore, useful for the study of income disparities.

Malee is only one of the several closest points on the convex hull when the model projects onto the query point Anong. There are more relevant points on the \hull in other sub-spaces of the domain. From the perspective of the Human Resources (HR) personnel, we want to disregard categorical variables (such as the country of origin, marital status, and gender) that should not be considered in determining salary. So, we relax the variable of the native country. Two samples now represent the closest point: a convex combination of two unmarried Vietnamese with weighted age of 26, with less education, and working a 60-hour week. Anong is much older and experienced, better educated, and works only 50 hours per week. If we present the result to a social scientist, the model would be criticized for typecasting Anong as a hard-working immigrant who focused on working until old age and had no spouse and no child. However, in the logic of a model, this projection might be reasonable: it extrapolates on a few continuous variables in the dimension of age, years of education, and working hours per week. Again, the extrapolation is significant.

Suppose the HR wants to relax the country of origin further to include both citizens and immigrants. In that case, the model finds the closest point represented by a group of single white women around the same age as Anong, who works less per week, averaging 41 hours per hour. What if the model also relaxes the categorical differences between man and woman (because gender really should not be considered in making a salary-related decision, although in reality, it is unfortunately not often the case). This time, the model extrapolates based on a group that averages at 42-year-old, 51 percent white, and 49 percent black male, who works 7 hours less per week than Anong. In these cases, the model extrapolates in the dimension of working hours per week. Anong is not a singular case. We find many query points requiring the model to extrapolate either in continuous features or both continuous and categorical features.

In social norms, extrapolating in the dimensions of categorical and continuous variables (gender, country of origin, and age) is shockingly excessive. The statistically significant adjustments are predetermined not by the bias embedded in the training set but by the fact that models have to extrapolate. We are not arguing that the bias in the datasets does not matter - they matter a lot. But here, we are concerned with the explainability issue, which we tackle with the tool to analyze the extent of extrapolation. The problem is not that we do not understand a model trained on this dataset. In fact, many have published on the bias in the models trained on this dataset \citep{besse2021survey}. The more persistent issue is that models have been extrapolating salary recommendations without informing the users and policymakers. This case shows what we understood the least, and the model does the most in many cases of extrapolation.

\subsection{The Second Case: FICO dataset}

The FICO dataset is a real-world financial model for determining the risk associated with traders who sell and buy goods, currency, and stock. A model trained on this dataset would predict the risk of traders. This dataset was the subject of challenge/competition at the 2018 NeurIPS conference \citep{fico2018challenge}. The best accuracy achieved on this dataset was by a simple model as opposed to a black-box model such as neural networks \citep{chen2018interpretable}. This dataset does not have any demographic information about individuals. Instead, all the 23 features are about financial transactions. This feature helps cut the weeds around models making mistakes in categorical variables, which we saw in the preceding case study about Anong. Moreover, we specifically chose this dataset because it was the subject of a challenge and further studies \citep{rudin2019why}. As a result, the dataset is relatively well-known in the AI research community. This study shows that extrapolation is still frequent and significant even for datasets locating a highly specific topic, with only continuous variables and no demographic information.

\begin{figure}[h]
 \centering
 \includegraphics[width=.5\linewidth]{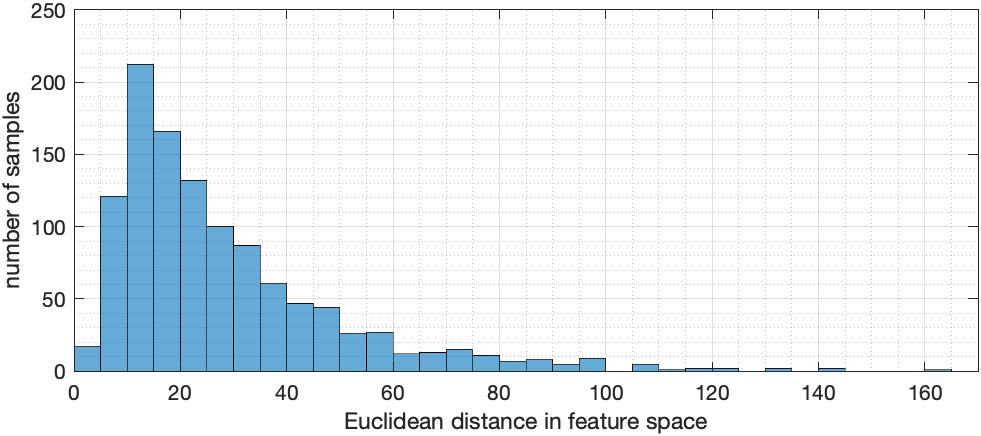}
 \caption{Distribution of distance to convex hull of training set for testing samples of FICO dataset that are outside the \hull. 54\% of samples for this dataset are outside the \hull.}
 \label{fig:fico_dist}
\end{figure}

Amongst the testing samples (traders), approximately half (54 percent) of them are outside the convex hull. Of the 54 percent, the histogram in Figure~\ref{fig:fico_dist} shows the distribution of distance to the convex hull. Note that in this dataset, we do not have any categorical variables. All variables are continuous, such as those denoting trade frequency, delinquency, and trade open time. 

Let us now pick a random trader at the average distance to the hull, which we show as testing point \#7 in Table~\ref{tbl:fico_sample}. Comparing the query point and its projection to the convex hull, we see that the model extrapolates moderately in the dimensions related to delinquency (percentage of trades never delinquent, and months since most recent delinquency). The extrapolation in the dimensions related to delinquency might seem very significant. However, if we look at other variables crucial for understanding the risk associated with a trader, we see a different story. As shown in Table~\ref{tbl:fico_sample}, we see that the model has to extrapolate much more significantly in several key dimensions, such as those related to trading frequency (e.g. the months since the trader's oldest trade is 30 months less and the trader's total number of credit accounts twice as few) and the amount of installment transactions (e.g. net fraction installment burden).

\begin{table}[h]
\setlength{\tabcolsep}{5pt}
\caption{Information for testing \#7 in the FICO dataset, and its projection to the convex hull of training set. Extrapolation is significant even for a typical sample in the dataset.}
\label{tbl:fico_sample}
\begin{center}
\begin{footnotesize}
\begin{sc}
\begin{tabular}{  p{5.5cm}  p{2cm}  p{2.5cm} }
\toprule
Variable Description & testing point \#7  & its projection to \hull   \\
\midrule
Consolidated version of risk markers   & 63  &64  \\
\midrule
Months Since Oldest Trade Open  &  54  & 84 \\
\midrule
Months Since Most Recent Trade Open  &  25  & 22\\
\midrule
Average Months in File  &  40   & 46\\ 
\midrule
Number of Satisfactory Trades   &  2   &  2\\
\midrule
Number of Trades 60+ Ever  & 0   & 0\\
\midrule
Number of Trades 90+ Ever  & 0   &  0\\
\midrule
Percentage of Trades Never Delinquent  & 50   & 52\\
\midrule
Months Since Most Recent Delinquency  & 22   & 21\\
\midrule
Max Delinquency/Public Records in the Last 12 Months  &  6   & 6\\
\midrule
Max Delinquency Ever  & 6 & 6\\
\midrule
Number of Total Trades (total number of credit accounts)  &  2    & 4\\
\midrule
Number of Trades Open in the Last 12 Months  &  0    & 1\\
\midrule
Percentage of Installment Trades  & 100   &  97\\
\midrule
Months Since Most Recent Inquiry (excluding last 7 days)  & -7   & -5\\
\midrule
Number of Inquiries in the Last 6 Months  & 4   & 4 \\
\midrule
Number of Inquiries in the Last 6 Months (excluding last 7 days)  & 4   & 4 \\
\midrule
Net Fraction Revolving Burden  & -8   & -8 \\
\midrule
Net Fraction Installment Burden  & -8   & 12 \\
\midrule
Number of Revolving Trades with Balance  & -8   & -8 \\
\midrule
Number of Installment Trades with Balance  & 1   & 1 \\
\midrule
Number of Bank/National Trades with High Utilization Ratio  & -8    & -8 \\
\midrule
Percentage of Trades with Balance  &  100   & 100 \\
\bottomrule
\end{tabular}
\end{sc}
\end{footnotesize}
\end{center}
\end{table}

If the model needs to extrapolate significantly about trader \#7, which is the average distance to the hull, those testing samples further away from the hull would logically require more significant extrapolation. The FICO case complements the Adult Income Dataset in showing the extent and the frequency of extrapolation in highly influential social applications of AI.

\subsection{The Third Case: Veterans Aging Cohort Study}

We now consider a very large real-world dataset containing 1,048,575 samples of electronic health record (EHR). This dataset is part of the Veterans Aging Cohort Study (VACS) \citep{justice2006veterans} at the veterans healthcare system. Dataset includes 23 features including demographic information, such as race, gender, and age, as well as, medical features related to blood tests, and whether the patient has Hepatitis C or not. All patients in this dataset are HIV+ and they have been on Highly Active Anti-retroviral Therapy (HAART) for more than a year.

Having more than a million samples provides the chance to investigate how extrapolation may occur by models used in the field of medicine. Indeed VACS study has been the subject of research from different perspectives, and predictive models have been developed based on this data \citep{tate2019improved}.

Using a 5-fold cross validation, we divide the samples into a training and testing set, where testing set contains 209,715 samples, representing 20\% of samples in the dataset. The rest of 838,860 samples form the training set. Analyzing the geometry of data reveals that approximately 15\% of testing samples are considerably outside the convex hull of training set. Average distance to the hull, for samples that are outside \hull, is approximately 175, measured in 2-norm in the feature space. Standard deviation of this distance is about 160. Let us pick a sample with distance 189 from the hull. Of the samples that are outside the hull, this sample is relatively typical as its distance is close enough to the average distance. Table~\ref{tbl:vacs_sample} shows the information for this sample, and also its projection to the convex hull of training set.


\begin{table}[h]
\caption{Features about patients considered in our analysis}
\label{tbl:vacs_sample}
\begin{center}
\begin{small}
\begin{sc}
\begin{tabular}{c c c c}
\toprule
Feature  & Value for X  & Closes point to it on the \hull \\
\midrule
Age & 59  &  54 \\
Race & Black & Black \\
Gender & Male & Male \\
CD4 count & 204 & 296 \\
Albumin & 3.3 & 4.3 \\
Alanine Aminotransferase & 45 & 95 \\
Aspartate Aminotransferase & 63 & 95 \\
Creatinine & 10.9 & 1.1 \\
Hemoglobin & 10.3 & 11.4 \\
Platelet count & 133 & 256 \\
White Blood Cell count & 2.3 & 2 \\
Body Mass Index & 15.6 & 19.9 \\
Number of days between visits & 63 & 26 \\
Time on Highly active anti-retroviral therapy & 15 years & 2.3 years \\
Fibrosis-4 & 4.17  & 2.06 \\
Estimated Glomerular Filtration Rate & 5 & 88 \\
Viral Load & 1.3 & 1.88 \\
Hepatitis C & Yes &  Yes \\
\bottomrule
\end{tabular}
\end{sc}
\end{small}
\end{center}
\end{table}

Many of the features for this patient, whom we call Bob, are noticeably far-off the closest point on the convex hull. Bob is 59 of age and underweight (his Body Mass Index is less than 16). For the sake of illustration, we describe the closest point on the convex hull in a figurative way as a 54-year-old man whose information is provided in the far right column in Table~\ref{tbl:vacs_sample}. This younger man, which is a convex combination of many samples in the training set, has BMI of 19.9 which can be considered healthy for this age. If the model generates health-related predictions for the underweight Bob, it will have to extrapolate based on younger and regular weight patients in the training set. If we look at other variables, we see further more severe implications of extrapolations. In medical terms, an estimated glomerular filtration rate (EGFR) below 60 would mean kidney disease for any given person, and Bob has an EGFR of only 5. Creatinine, another indicator of kidney function, is significantly higher than his closest projection on the convex hull. Third, a Fibrosis-4 index higher than 3.25 indicates significant liver fibrosis, and Bob falls into that category. Bob and his projection to the convex hull would be treated as two entirely different cases to medical experts. Yet, a model trained on our rather large training set has no choice but to extrapolate from less ill patients to estimate the health condition of a much sicker patient like Bob. We can logically conclude that the extent of extrapolation is notable. And there are many other patients in the dataset falling outside the convex hull of training. Any given model would need to extrapolate for them.

To promote transparency and accountability, a model should inform the clinicians and physicians whenever it extrapolates, i.e., let the users of the model know when a given sample is not similar to the samples it has seen before. A trained model may make a correct or incorrect prediction for Bob. Either way, it is reasonable that a human expert should be notified about the extrapolation performed by the model, and review the recommendation/evaluation of the model made for such patient.

\section{Policy implications and future work} \label{sec:policy}

Many different applications affecting humans use AI models. There have been various calls for regulating AI models to make them explainable. Notably, the European Union has drafted and passed the General Data Protection Regulation (GDPR), which requires decisions made by automated systems to be accompanied by clear explanations. Our work, in this paper, suggests that extrapolation should be an inherent part of such explanations.

The postulation that a machine learning model can automatically learn from the data and generalize well seems overly optimistic. Frequently, it has been reported that AI models may exercise various sorts of unacceptable behaviors and biases \citep{oneil2016weapons,angwin2016machine}. Sometimes such behaviors are rooted in the bias in the data. AI models may even exacerbate those biases or develop new preferences \citep{roosli2021bias}. But, what if the model has to extrapolate to significant degrees for a considerable portion of its testing samples? The crucial insight here is not that datasets contain bias, but rather, algorithms can freely and significantly extrapolate in many dimensions.

From the public policy perspective, our study shows the benefit of creating AI pipelines that automatically report to relevant entities whenever they extrapolate significantly. In high-stakes decision-making, it might be sensible to impose regulations requiring AI models' extrapolated decisions/predictions to be sent for human expert review.

\subsection{Privacy issues}

Reporting that a testing sample falls outside the convex hull and that a model needs to extrapolate to process it, in our view, is crucial information that should be shared with the entities affected by the output of AI models. However, other questions arise. In specific applications of machine learning, it may not be possible to share information about training samples.

Consider, for example, an AI model that recommends clinical decisions based on patient data. If the model has to extrapolate for a particular patient, it would be useful to let the physician know that the model has extrapolated. Suppose physicians have the necessary authorization to access other patients' data. In that case, it may be useful for them to see the information about other relevant patients, i.e., the convex combination of samples that form the $\mathcal{P}^h(x,\mathcal{H}^{tr},\Omega)$ for patient $x$. But, if we want to report the extrapolation to the patient him/herself, it might be best practice to just report that the model had to extrapolate, and perhaps, accompany that with additional information that extrapolation happened in which dimensions, e.g., age and BMI. But, it may not be proper nor legal to share which training samples were closest to the patient's data, as it may violate the privacy of other patients. 

Our formulated auditing procedures require access to the training samples, but due to privacy issues, training sets could be protected from auditing procedures. In such cases, one approach could be to approximate the convex hull of training set using a different set of points other than the original set. There are several efficient methods available for approximating convex hulls with theoretical bounds on the maximum difference between the approximated convex hull and the original one \citep{blum2018approximate,blum2019sparse}. Approximation can also address the first privacy concern discussed above. 

The extent to which an AI model should report its extrapolation is an important topic that can be the subject of further studies on privacy and transparency.


\subsection{Implications for model selection}

We also observe, empirically, that extrapolation is less significant for learning tasks where deep networks have not shown an advantage over simple models. On the other hand, the extent of extrapolation is larger for learning tasks where deep networks have an advantage. This suggests that the geometry of data, and specifically, the relative extent of extrapolation, might be a useful guide for model selection. When the extent of extrapolation is small, simple models may be more appealing, especially if they achieve an accuracy as good as deep networks, because simple models are computationally less expensive and easier to interpret. When the extent of extrapolation is more significant, more sophisticated models such as deep networks might be more likely to achieve better accuracy. These implications may be studied in future studies in the context of model selection literature.


\section{Conclusions} \label{sec:conclusion}
 
This paper promotes transparency in AI by focusing on the neglected factor of extrapolation. Countering the common belief that AI models do not extrapolate much, we show the frequency and the non-negligible extrapolation through geometric analyses of several social applications of AI related to salary, trading, and health care.

Having reviewed the literature on accountability and transparency, we present our auditing approach to clear the current roadblocks. Most research published on these datasets has discussed bias inherent in the training set, but few offer ways to audit how much the models extrapolate. This paper proposes a practical solution: when extrapolation happens (which has been happening so far without notice), the model should have a self-auditing system to report the extent of extrapolation. If extrapolation is statistically excessive, the model should generate a report indicating to what extent and in which features has the model extrapolated. 

We illustrate with examples how socially consequential extrapolation can be and how to add a robust auditing step to the AI pipeline. The extrapolations we have examined can happen in multiple dimensions - both continuous features and categorical features. A testing sample outside the convex hull requires the model to project in both categorical and continuous variables. Even in cases where categorical variables are not involved, extrapolation is still non-negligible. 

We use a systematic method of geometric analysis in developing an auditing approach that can be applied automatically to any dataset/model. Our systematic approach is specifically useful in considering domains with a combination of categorical, integer, and continuous variables. We suggest that models be audited with respect to extrapolation, and reporting of extrapolation becomes part of regulations such as GDPR.

\section*{Acknowledgements}

R.Y. thanks Cynthia Rudin for a helpful comment on earlier version of this work. R.Y. was supported by a fellowship from the Department of Veterans Affairs. The views expressed in this manuscript are those of the authors and do not necessarily reflect the position or policy of the Department of Veterans Affairs or the United States government.

\bibliographystyle{ACM-Reference-Format}
\bibliography{refs,refs_cog}

\end{document}